# MEDFORM: A Foundation Model for Contrastive Learning of CT Imaging and Clinical Numeric Data in Multi-Cancer Analysis

**Daeun Jung, Jaehyeok Jang, Sooyoung Jang, Yu Rang Park**

Department of Biomedical Systems Informatics, Yonsei University College of Medicine

alleun1110@yonsei.ac.kr, jhyeok@yush.ac, jan9sy89@yuhs.ac, yurangpark@yuhs.ac

## Abstract

*Computed tomography (CT) and clinical numeric data are essential modalities for cancer evaluation, but building large-scale multimodal training datasets for developing medical foundation models remains challenging due to the structural complexity of multi-slice CT data and high cost of expert annotation. In this study, we propose MEDFORM, a multimodal pre-training strategy that guides CT image representation learning using complementary information from clinical data for medical foundation model development. MEDFORM efficiently processes CT slice through multiple instance learning (MIL) and adopts a dual pre-training strategy: first pretraining the CT slice feature extractor using SimCLR-based self-supervised learning, then aligning CT and clinical modalities through cross-modal contrastive learning. Our model was pre-trained on three different cancer types: lung cancer (141,171 slices), breast cancer (8,100 slices), colorectal cancer (10,393 slices). The experimental results demonstrated that this dual pre-training strategy improves cancer classification performance and maintains robust performance in few-shot learning scenarios. Code available at https://github.com/DigitalHealthcareLab/25MultiModalFoundationModel.git*

## 1. Introduction

Computed tomography (CT) is an indispensable diagnostic tool for cancer offering detailed insights into anatomical changes and potential abnormalities. CT imaging is essential for detecting and evaluating various cancers. With the advance in deep learning models such as convolutional neural networks (CNNs), analysis of CT imaging has significantly enhanced its clinical application such as disease classification and prognosis evaluation. However, acquiring large amounts of expert-labeled data poses significant challenges due to high-cost and professional resources.

Self-supervised learning (SSL) with contrastive learning enables models to identify and differentiate patterns in images by comparing similar and dissimilar image leveraging unlabeled data alone [1]. Through this approach, the model can capture underlying anatomical features from CT imaging and perform downstream tasks effectively. Also the weakly supervised learning via multiple instance learning (MIL) achieved successful performance in medical images such as computational pathology [2]. CT slice images the analysis of CT imaging performs patient-level evaluation beyond individual CT slice evaluation. So, MIL can be considered as a suitable approach for CT imaging as it does not require slice-level expert annotation.

In addition to structural evaluation using CT imaging, clinical data—including demographics and disease history—plays a crucial role in decision-making as clinical data are known to correlate with tumor progression, patient prognosis. Deep learning models have been developed to effectively handle tabular data, demonstrating promising performance with interpretability and further enhance their ability to learn meaningful representations with SSL [3].

Integrating these modalities, CT imaging and clinical data, enables a comprehensive analysis based on both structural and systemic information. Contrastive learning can be adopted for multimodal models by aligning embeddings from CT imaging and clinical data. This allows the model to leverage the relationships between different modalities while preserving characteristics from each modality, CT imaging and clinical tabular data.

In this study, we developed multimodal model with contrastive learning between CT imaging and clinical data for multiple cancer data to align embeddings from pretrained models of CT imaging and clinical data. We trained each unimodal models on large-scale data of CT imaging with contrastive learning and clinical tabular data using encoder-decoder to extract representations of data. Then, multimodal models with multimodal contrastive to the relationship of extracted representations of each modality. Our contributions are as follows

- Pretrained unimodal models of CT imaging and clinical data with self-supervised learning
- Multimodal contrastive learning using embedding from pretrained models
- Few-shot learning for cancer evaluation (staging and classification)

## 2. Related work

### 2.1 Deep learning in medical imaging

The complexity and heterogeneity of medical data necessitates a multimodal approach, as different modalities provide complementary information crucial for comprehensive clinical understanding [4]. A recent study has demonstrated that combining histopathology images and clinical staging data for endometrial cancer prognosis outperformed conventional approaches and improved treatment personalization [5]. In this context, numerous studies are developing medical foundation models capable of learning from diverse data types [6-9]. As the field requires larger datasets to improve model performance, the challenges of comprehensive data labeling have increased the use of self-supervised learning approaches, enabling models to learn meaningful representations from unlabeled multimodal medical data [5, 7].

### 2.2 Multiple instance learning in medical imaging

Multiple instance learning (MIL) is a learning approach where data is organized into bags containing multiple instances, with labels assigned at the bag level rather than to individual instances, thus reducing the burden of data annotation while improving computational efficiency [10, 11]. The structure of medical imaging data, where multiple images (instances) belong to a single patient (bag), matches this MIL approach, providing an efficient framework for patient-level analysis [12]. While MIL has been widely adopted in pathology imaging analysis, it has also been broadly applied to CT imaging for various clinical applications, including the detection of intracranial hemorrhage and various pulmonary conditions [2, 13-16].

### 2.3 Contrastive learning

SSL method with contrastive learning utilizes contrastive objectives that maximize mutual information between positive pair and minimized it for negative pair. SimCLR [1] introduced a simple framework leveraging augmented views of the same image as positive pair. It demonstrated improved performance over previous SSL methods on ImageNet. MoCo [17] proposed a momentum-based contrastive learning approach, utilizing a large dictionary for efficient training with smaller batch sizes. The representations learned from MoCo have shown improved performance in downstream task such as object detection and segmentation.

### 2.4 Cross-modal contrastive learning

Cross-modal learning makes it possible to seamlessly integrate various data types by extracting relevant information from one modality (e.g. image) based on queries from another modality (e.g. clinical data, genetic data) [18-22]. Most representatively, CLIP, jointly trains an image encoder and a text encoder to predict the correct pairings of a batch of (image, text) examples [23]. During training, CLIP maximizes the cosine similarity between embeddings of matched pairs while minimizing similarity between unmatched pairs in a shared latent space [23]. This approach has been adapted to medical domains such as combining MRI imaging data with clinical tabular features [24], and integrating chest X-ray images with corresponding radiology reports data through multimodal contrastive learning [25]. However, limited studies have explored cross-modal contrastive learning between computational tomography images and clinical numeric data, motivating us to investigate the relation between these two complementary data types

## 3. Method

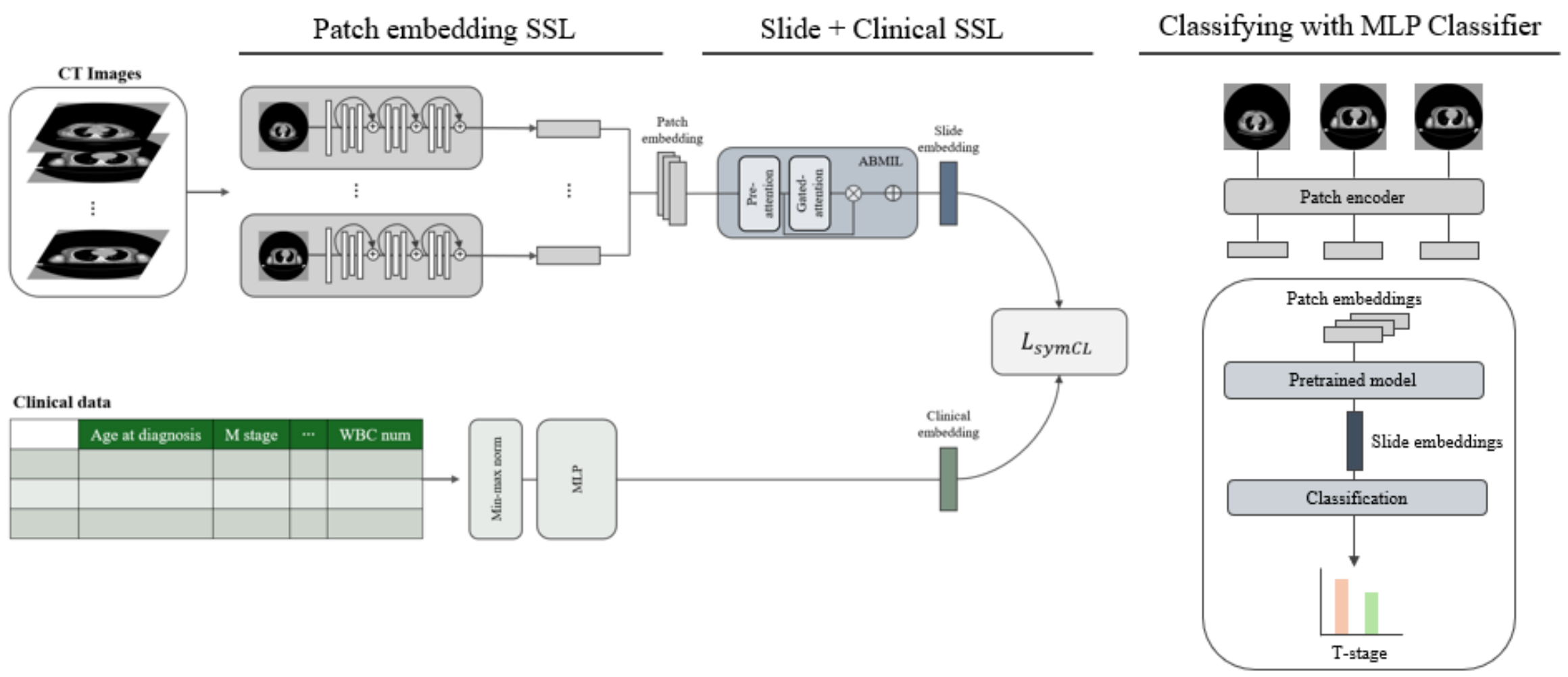

**Figure 1. Overview of MEDFORM (S+C) pre-training**

We adapted TANGLE [22], a transcriptomics-guided representation learning framework, for joint pre-training of CT images and clinical data through contrastive learning. The proposed framework consists of three main components: (1) a vision encoder that encodes slice-level embeddings from sequential CT slices and is pooled by a module to aggregate them into a unified CT representation; (2) a clinical data encoder that generates embedding using clinical numeric features; and (3) a multimodal alignment module that learns to project both CT and clinical representations into a shared semantic space.

### 3.1 Slice-based volumetric CT encoder

We adopted the Multiple Instance Learning (MIL) paradigm [10, 11, 26], traditionally applied to whole slide images (WSI), to effectively process and learn from volumetric CT data. Through parallel sectioning along the transverse plane, the 3D CT volumes were converted into successive 2D axial slices. To create a single CT representation, the sliced 2D CT image embeddings from each slice were then aggregated through pooling operation.

**Preprocessing CT images:** Volumetric CT data from DICOM files were successively sorted along the z-axis for data pre-processing in order to maintain proper slice alignment. To improve model performance, we applied circular mask with a size of $512 \times 512mm$ during CT preprocessing [27]. This allowed us to effectively extract the Region of Interest (ROI) and remove unnecessary background information from the CT images. The CT images were resampled to uniform spatial resolution of $1.0 \times 0.8 \times 0.8mm$, followed by cancer-specific Hounsfield Unit (HU) windowing to enhance the visibility of critical structures such as tumor, adipose, and glandular tissues. Subsequently, the HU values were normalized to the range [0,1].

**Pre-trained CT slice encoding :** For slice-level encoding, we trained ResNet50 from scratch using SimCLR [1] framework, denoted as CT-SimCLR, on a dataset comprising 141,171 NSCLC slices, 8,100 breast cancer slices, and 10,323 colorectal cancer images. CT-SimCLR represents a pioneering pretrained model specifically designed for 2D CT-slice representations. We let the resulting CT slice embeddings of the $i$-th slide $X_i$ as $S_i \in R^{N_H \times d_H}$, where $N_H$ is the number of slices volumetric CT scan and $d_H$ denotes their embedding dimension.

**Volumetric Feature Aggregation via MIL**: We formulate a function $f(S_i): R^{N_H \times d_H} \rightarrow R^d$ that maps the set of CT slice embeddings into comprehensive volumetric representation $s_i \in R^d$. To implement this mapping function, we used an attention-based Multiple Instance Learning framework(ABMIL) [11] that dynamically allocates importance weights to individual slices using an advanced slice-level attention mechanism. These weighted slice embeddings are then aggregated through a pooling operation to generate global CT representation.

### 3.2 Clinical numeric data encoder

Given a set of clinical datasets, denoted as $t_i \in R^{N_C}$, where $N_C$ represents the dimension of clinical features, we employ a multilayer perceptron (MLP) for encoding, represented as $\varphi(\cdot)$. Specifically, for clinical data encoder, we train a 3-layer MLP $\varphi(t_i) : R^{N_C} \rightarrow R^d$to project the clinical representation $t \in R^{N_C}$ to an expression embedding $c_i \in R^d$, enabling alignment within the same latent space as CT image embeddings.

### 3.3 Multimodal alignment

**Pre-training contrastive learning alignment:** We align CT image and clinical data encoders into a unified embedding space through a symmetric cross-modal contrastive learning objective. This approach, commonly used in vision-language pre-training models , allows effective multi-modal representation learning [28]. Formally, for each batch, we define paired embeddings $(s_i, c_i)$, where $s_i$ represents the $i$-th CT slice embedding and $c_i$ denotes its corresponding clinical data expression. To align these cross-modal representations is shared embedding space, we minimize the following symmetric contrastive loss function:

$$L_{SymCL} = -\frac{1}{2M}\sum_{i=1}^{M} log \frac{\exp(\tau s_i^T c_i)}{\sum_{j=1}^{M}\exp(\tau s_i^T c_j)} - \frac{1}{2M}\sum_{j=1}^{M} log \frac{\exp(\tau c_j^T s_j)}{\sum_{j=1}^{M}\exp(\tau c_j^T s_i)} \quad (1)$$

The loss function is composed of two terms: a CT slice-to-clinical contrastive loss and a clinical-to-CT slice contrastive loss. Each term functions as a contrastive sample by treating other samples in the batch as negative pairs and optimizing the normalized dot product similarity between paired embeddings.

During inference, each CT slide is processed through the pre-trained vision encoder to extract CT slice-level embeddings, which are then aggregated the MIL module to generate a volumetric representation aligned with the clinical embedding space. These aligned representations are then used for downstream tasks, such as classification and few-shot learning using linear probing.

To compare our model's performance, we used a model consisting of (1) a vision encoder and (2) a clinical data encoder using TabNet [3]. For the vision encoder, CT slice embeddings were first extracted through CT-SimCLR and then aggregated with a pooling module to generate volumetric representations. The feature vectors from two modalities were concatenated and processed through multi-layer perceptron for final prediction.

## 4. Experiments and results

### 4.1 Dataset

We collected CT images and corresponding clinical data in three distinct cancer types – non-small cell lung cancer (NSCLC), breast cancer, and colorectal cancer-through The Cancer Imaging Archive (TCIA) portal. To guarantee data consistency, we only used CT scans obtained using standardized imaging techniques and under uniform imaging protocols for each cancer type.

Non-small cell lung cancer (NSCLC): For NSCLC, we collected approximately 150,000 slices of CT images from four cohort: NSCLC Radiogenomics, NSCLC Radiomics, COVID-19-AR, MIDRC-RICORD-1B. Among these, 101,117 slices from NSCLC Radiomics, COVID-19-AR, MIDRC-RICORD-1B were used for CT-SimCLR pre-training and the rest from NSCLC Radiogenomics were used in training multimodal alignment training and as test set.

Breast cancer: We collected CT images totaling 10,050 slices of CT image from three cohorts: ACRIN-FLT-Breast, NSCLC Radiogenomics, and Lung-PET-CT-Dx. Only thorax part was used in NSCLC Radiogenomics and Lung PET-CT Dx data. NSCLC Radiogenomics and Lung PET-CT-Dx datasets were fully used to pretrain CT-SimCLR, and part of ACRIN-FLT-Breast was used in CT-SimCLR pretraining, and rest of it was used in training multimodal alignment and as test set.

Colorectal cancer: For colorectal cancer, we collected 52,917 slices from Colorectal Liver Metastatus cohort and used 10,393 slices to train CT-SimCLR, and the rest were used for training and rest of it was used in training multimodal alignment and as test set.

**Table 1. Performance of unimodal model, feature concatenation model, and contrastive learning model for multiple cancer types.** AUROC = Area Under the Receiver Operating Characteristic curve, ACC = Accuracy

| Cancer type | Cohort | Task | Metric | Unimodal | Multimodal (Feature concatenation) | Multimodal (Contrastive learning) |
|---|---|---|---|---|---|---|
| Breast Cancer | ACRIN | T-stage | AUROC | 0.6390 (0.1212) | 0.6300 (0.1831) | **0.7042 (0.1770)** |
| | | | ACC | 0.5333 (0.1165) | 0.5000 (0.0791) | **0.5393 (0.0510)** |
| Colorectal cancer | TCIA | Bilobar disease | AUROC | 0.4825 (0.0755) | 0.5151 (0.0149) | **0.5336 (0.1384)** |
| | | | ACC | 0.5078 (0.0635) | 0.5078 (0.0635) | **0.6583 (0.0283)** |
| Non-Small Cell Lung Cancer | TCIA | Histologic stage | AUROC | 0.6000 (0.1290) | 0.6086 (0.1520) | **0.6658 (0.1092)** |
| | | | ACC | 0.7788 (0.1020) | **0.8420 (0.0000)** | 0.8208 (0.0234) |

## 4.2 Cancer classification using multimodal contrastive learning

We evaluate our proposed framework for cancer classification across multiple cancer datasets, including breast cancer (ACRIN), colorectal cancer (TCIA), and non-small cell lung cancer (TCIA). The framework was tested on different classification tasks specific to each cancer type: breast cancer T-stage classification, colorectal cancer bilobar disease assessment, and non-small cell lung cancer histologic staging. We benchmark the performance using three different approaches: (1) a unimodal ResNet-50 model for CT images pretrained with SimCLR, (2) a feature concatenation model using the same pretrained model for CT with tabular data, and (3) our proposed multimodal contrastive learning model.

As shown in Table 1, our multimodal contrastive learning approach consistently outperforms all baselines across the three cancer datasets. For breast cancer T-stage classification, our model achieves an AUROC of 0.7042 and ACC of 0.5393, showing substantial improvement over both the unimodal (AUROC: 0.6390) and feature concatenation (AUROC: 0.6300) approaches. Similar performance gains are observed in colorectal cancer bilobar disease classification and non-small cell lung cancer histologic stage classification, where the proposed multimodal contrastive learning approach achieves AUROC scores of 0.5336 and 0.6558, respectively.

**Table 2. Few-shot learning performance of multimodal contrastive learning model for multiple cancer types.** AUROC = Area Under the Receiver Operating Characteristic curve, ACC = Accuracy

| Cancer type | Cohort | Task | Model | Metric | k=1 | k=5 | k=10 |
|---|---|---|---|---|---|---|---|
| Breast Cancer | ACRIN | T-stage | TANGLE | AUROC | 0.508 (0.232) | 0.659 (0.068) | 0.687 (0.129) |
| | | | | ACC | 0.524 (0.144) | 0.628 (0.051) | 0.642 (0.091) |
| Non-Small Cell Lung Cancer | TCIA | Histologic stage | TANGLE | AUROC | 0.425 (0.142) | 0.523 (0.050) | 0.582 (0.129) |
| | | | | ACC | 0.458 (0.130) | 0.520 (0.093) | 0.630 (0.125) |

Additionally, we evaluated the few-shot learning performance of our proposed model across different cancer classification tasks. The model demonstrates robust performance even with limited training samples, as shown in Table 2. For breast cancer T-stage classification, with five shots (k=5), the model achieves an AUROC of 0.659 and accuracy of 0.628, demonstrating reasonable performance even with minimal training data. The performance further improves with additional samples, reaching an AUROC of 0.687 and accuracy of 0.642 at k=10.

## 5. Conclusion

In this paper, we introduced a novel contrastive learning framework for effectively integrating CT images and clinical numerical features for cancer classification. Our approach was evaluated across multiple cancer datasets and various classification tasks, consistently demonstrating superior performance compared to baseline models.

A notable advantage of our approach is its efficient representation of medical imaging data through MIL [10, 12]. Medical images, particularly CT and MRI scans, often generate multiple high-resolution images per patient, which can be challenging to process directly. While MIL has been widely adopted in medical imaging analysis and CT applications due to these characteristics, it is inherently optimized for binary classification tasks [14-16]. Recent work by Jaume et al. explored combining MIL with contrastive learning for pathology image analysis, integrating whole-slide images with gene expression data[22]. We apply this joint approach of MIL and contrastive learning to CT imaging - where MIL effectively organizes multiple images as instances within patient-level bags, and contrastive learning enables robust feature learning across different modalities. In our study, this integration results in both computational efficiency and improved performance.

The results clearly show that multimodal learning contributes significantly to performance improvement compared to unimodal approaches. While unimodal models are limited by their single information source, multimodal approaches can learn richer, more comprehensive representations of both imaging and clinical characteristics [4, 29]. Furthermore, our results suggest that contrastive learning enables more effective integration of multiple modalities compared to simple feature concatenation, leading to improved model performance through better alignment of CT imaging and clinical measurement spaces [23].

Future work includes extending our framework to incorporate additional modalities such as genomic data and chest X-rays to enhance the representation learning. Additionally, investigating the model explainability could provide deeper understanding of how the model integrates different types of medical data.